\documentclass[10pt,twocolumn,letterpaper]{article}

\usepackage{wacv}              % To produce the CAMERA-READY version
\usepackage{graphicx}
\usepackage{amsmath}
\usepackage{amssymb}
\usepackage{booktabs}

\usepackage{multirow}
\usepackage{pgfplots}

\usepackage[pagebackref,breaklinks,colorlinks]{hyperref}
\usepackage{hyperref}

\usepackage[capitalize]{cleveref}
\crefname{section}{Sec.}{Secs.}
\Crefname{section}{Section}{Sections}
\Crefname{section}{Section}{Sections}
\Crefname{table}{Table}{Tables}
\crefname{table}{Tab.}{Tabs.}

\def\wacvPaperID{227} % *** Enter the WACV Paper ID here
\def\confName{WACV}
\def\confYear{2024}

\begin{document}

\title{TransRadar: Adaptive-Directional Transformer for Real-Time Multi-View Radar Semantic Segmentation}
% Direction-Adaptive Transformer for Real-Time Radar Semantic Segmentation

\author{Yahia Dalbah
\qquad
Jean Lahoud
\qquad
Hisham Cholakkal\\
Mohamed Bin Zayed University of Artificial Intelligence (MBZUAI)
}
\maketitle

%%%%%%%%% ABSTRACT
\begin{abstract}
   % Scene understanding is an important aspect in enabling automotive driving technologies and allowing for higher standards of performance and safety. 
   Scene understanding plays an essential role in enabling autonomous driving and maintaining high standards of performance and safety.
   % Cameras and laser scanners (LiDARs) have been the most common approaches to tackling this task, with radars being a less popular approach. 
   To address this task, cameras and laser scanners (LiDARs) have been the most commonly used sensors, with radars being less popular.
   Despite that, radars remain low-cost, information-dense, and fast-sensing techniques that are resistant to adverse weather conditions. 
   % Multiple works tackling scene semantic segmentation have previously risen in the literature, but are still limited due to the lack of large-scale annotated datasets and research in radar-frequency semantic segmentation. 
   While multiple works have been previously presented for radar-based scene semantic segmentation, the nature of the radar data still poses a challenge due to the inherent noise and sparsity, as well as the disproportionate foreground and background.
   In this work, we propose a novel approach to the semantic segmentation of radar scenes using a multi-input fusion of radar data through a novel architecture and loss functions that are tailored to tackle the drawbacks of radar perception. 
   Our novel architecture includes an efficient attention block that adaptively captures important feature information. 
   Our method, TransRadar, outperforms state-of-the-art methods on the CARRADA \cite{carrada} and RADIal \cite{radial} datasets while having smaller model sizes. 
   %We achieve a mIoU score of \textbf{63.9\%} and \textbf{47.5\%} for the semantic segmentation task in the CARRADA dataset on the Range-Doppler and Range-Angle maps, respectively. We also achieve a mIoU of \textbf{81.1\%} on the RADIal dataset's semantic segmentation task and an AP of \textbf{97.3\%} and AR of \textbf{98.4\%} on the object detection task. 
   \href{https://github.com/YahiDar/TransRadar}{https://github.com/YahiDar/TransRadar}
\end{abstract}

%%%%%%%%% BODY TEXT
%%%%%%%%% BODY TEXT
\section{Introduction}
% Radar sensing has been regularly used for most of the instrumented deterministic measurements, such as speed, blind spot detection, and automatic cruise control. This has been attributed to their relatively lower cost, lower processing time, and ability to report the velocity of objects. However, LiDAR sensors have risen in popularity as the main automotive perception tool for autonomous driving due to their relatively higher resolution and ability to generate detailed point-cloud data. LiDARs higher resolution granted them popularity in recent literature in the tasks of object detection and semantic segmentation. However, LiDARs suffer from a few drawbacks due to the shorter wavelength of their signals. LiDAR sensors are highly prone to errors, weather fluctuations, and occlusion with raindrops and/or dust \cite{9000872}. Furthermore, LiDAR signals' higher frequencies result in a rapid attenuation of their strength with respect to distance traveled, which results in a maximum range of operation of 100 to 200m.
Automotive systems rely on radar sensing for most of the tasks that require deterministic distance measurements, such as collision avoidance, blind spot detection, and adaptive cruise control. 
The prevalence of radar sensors in these tasks has been attributed to their relatively low cost, low processing time, and ability to measure the velocity of objects. 
On the other hand, LiDAR sensors have risen in popularity as the main automotive perception tool for autonomous driving due to their relatively higher resolution and ability to generate detailed point-cloud data.
This popularity is noticeable in recent literature, where LiDAR sensors are dominantly used in object detection and semantic segmentation tasks.
However, LiDAR sensors suffer from few drawbacks originating from the shorter wavelength of their signals. 
LiDAR sensors are highly prone to errors, weather fluctuations, and occlusion with raindrops and/or dust \cite{9000872}.
Moreover, LiDAR signals' higher frequencies result in a rapid attenuation of their strength with respect to distance traveled, which results in a maximum range of operation of 100 to 200m.

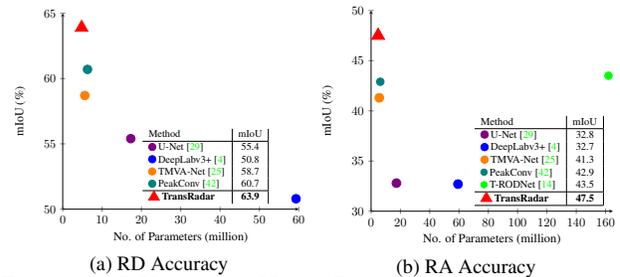
\begin{figure}[t!]
\centering
\resizebox{8.25cm}{!}{
\begin{minipage}{0.499\linewidth}
	\resizebox{\linewidth}{!}{%
		\begin{tikzpicture} 
		\begin{axis}[
		axis lines = left,
		% title={$y=a \cdot x + b$},  %xmin=10, xmax=100,
		ymin=50, ymax=65, 
		xmin=0, xmax=60,
		xlabel=No. of Parameters (million),
		ylabel= mIoU (\%),
		]
		\coordinate (legend) at (axis description cs:0.90,0.006);
		%%%% FPN
		
		%% Grid RCNN
		\addplot[only marks,
		mark=otimes*, violet,
		mark size=3.5pt
		]
		coordinates {
			(17.3,55.4)};\label{plot:U-Net}
		%% Cascade R-CNN
		\addplot[only marks,
		mark=otimes*, blue,
		mark size=3.5pt
		]
		coordinates {
			(59.3,50.8)};\label{plot:DeepLabv3}
        
  %       \addplot[only marks,
		% mark=otimes*, blue,
		% mark size=3.5pt
		% ]
		% coordinates {
		% 	(106.4,56.6)};\label{plot:RAMP-CNN}
		% %% Libra R-CNN 
		% \addplot[only marks,
		% mark=otimes*, cyan,
		% mark size=3.5pt
		% ]
		% coordinates {
		% 	(2.4,29.0)};\label{plot:MVNet} 
		%% Trident 
		\addplot[only marks,
		mark=otimes*,  orange,
		mark size=3.5pt
		]
		coordinates {
			(5.6,58.7)};\label{plot:TMVANet}
        \addplot[only marks,
		mark=otimes*,  teal,
		mark size=3.5pt
		]
		coordinates {
			(6.3,60.7)};\label{plot:PeakConv}
		%% Ours 
		\addplot[only marks,
		mark=triangle*, red,
		mark size=6pt
		]
		coordinates {
			(4.8,63.9)};\label{plot:ours}
		%% Grid RCNN plus
		
		\end{axis}
		\node[draw=none,fill=none,anchor= south east] at (legend){\resizebox{3.8cm}{!}{
				\begin{tabular}{l|c|c}
    
				Method  &mIoU\\ \hline

				\ref{plot:U-Net} U-Net~\cite{unet}   & 55.4   \\
    			\ref{plot:DeepLabv3} DeepLabv3+~\cite{deeplabv3+}   & 50.8   \\
				\ref{plot:TMVANet} TMVA-Net~\cite{2021multiview}   & 58.7  \\
				\ref{plot:PeakConv} PeakConv~\cite{zhang2023peakconv}   & 60.7  \\\hline
                
				\ref{plot:ours_RA} \textbf{TransRadar}  & \textbf{63.9}   \\
				\end{tabular}  }};
		\end{tikzpicture}  }
	\centering  \footnotesize (a) RD Accuracy \vspace{-0.2cm}
\end{minipage}
%\qquad
\begin{minipage}{.499\linewidth}
	%\resizebox{}{}{}
	\resizebox{\linewidth}{!}{%
		\begin{tikzpicture} 
		\begin{axis}[
		axis lines = left,
		% title={$y=a \cdot x + b$},  %xmin=10, xmax=100,
		ymin=30, ymax=50, 
		xmin=0, xmax=165,
		xlabel=No. of Parameters (million),
		ylabel= mIoU (\%),
		]
  \coordinate (legend) at (axis description cs:0.99,0.006);
		%%%%%% mask RCNN 

		%% Grid RCNN
		\addplot[only marks,
		mark=otimes*, violet,
		mark size=3.5pt
		]
		coordinates {
			(17.3,32.8)};\label{plot:U-Net_RA}
		%% Cascade R-CNN
		\addplot[only marks,
		mark=otimes*, blue,
		mark size=3.5pt
		]
		coordinates {
			(59.3,32.7)};\label{plot:DeepLabv3_RA}
        
  %       \addplot[only marks,
		% mark=otimes*, blue,
		% mark size=3.5pt
		% ]
		% coordinates {
		% 	(106.4,27.9)};\label{plot:RAMP-CNN_RA}
		%% Libra R-CNN 
		% \addplot[only marks,
		% mark=otimes*, cyan,
		% mark size=3.5pt
		% ]
		% coordinates {
		% 	(2.4,26.8)};\label{plot:MVNet_RA} 
		%% Trident 
		\addplot[only marks,
		mark=otimes*,  orange,
		mark size=3.5pt
		]
		coordinates {
			(5.6,41.3)};\label{plot:TMVANet_RA}
		%% Ours 
        \addplot[only marks,
		mark=otimes*, green,
		mark size=3.0pt
		]
		coordinates {
			(162,43.5)};\label{plot:T-RODNet_RA}
     \addplot[only marks,
		mark=otimes*, teal,
		mark size=3.0pt
		]
		coordinates {
			(6.3,42.9)};\label{plot:PeakConv_RA}
		\addplot[only marks,
		mark=triangle*, red,
		mark size=6pt
		]
		coordinates {
			(4.8,47.5)};\label{plot:ours_RA}
		%% Grid RCNN plus

		%% Ours

		\end{axis}
		\node[draw=none,fill=none,anchor= south east] at (legend){\resizebox{3.8cm}{!}{ \begin{tabular}{l|c|c}
				Method  &mIoU\\ \hline

				\ref{plot:U-Net} U-Net~\cite{unet}   & 32.8    \\
    			\ref{plot:DeepLabv3_RA} DeepLabv3+~\cite{deeplabv3+}   & 32.7    \\
				% \ref{plot:MVNet_RA} MVNet~\cite{2021multiview}   & 26.8 & 28.5   \\
				\ref{plot:TMVANet_RA} TMVA-Net~\cite{2021multiview}   & 41.3   \\
               \ref{plot:PeakConv_RA} PeakConv~\cite{zhang2023peakconv} & 42.9   \\
                \ref{plot:T-RODNet_RA} T-RODNet~\cite{TRODNet}   & 43.5    \\				\hline
                
				\ref{plot:ours_RA} \textbf{TransRadar}  & \textbf{47.5}   \\
    
				\hline
				\end{tabular}  }};
		\end{tikzpicture}}
	\centering  \footnotesize (b) RA Accuracy \vspace{-0.2cm}
\end{minipage}
}
\caption{mIoU scores vs No. of Parameters (millions) of state-of-the-art models in semantic segmentation on the CARRADA dataset. Our method, TransRadar, outperforms previous state-of-the-art methods in the semantic segmentation task with an mIoU of 63.9\% for RD maps and 47.5\% for RA maps.} \vspace{-0.2cm}
\label{fig:intro_fig}
\end{figure}

% # 
% Unlike LiDARs, frequency-modulated continuous wave radars operate in the millimeter wave band in which signals do not get significantly attenuated when faced with occlusions, allowing operation ranges of up to 3,000m.  Radar sensing has a long and solid history in deterministic instrumentation, and its information-rich nature results in it being a good data pipeline for computer vision tasks \cite{rodnet, raddet}. Radars however suffer from having poor resolution relative to LiDARs. Despite that, radars function in longer ranges and in adverse weather conditions more robustly than other commonly used sensing methods like cameras and LiDARs. Radar signals are rich in information as they contain Doppler information which contains the velocity of the objects. These signals can be processed to be used in an image-like pipeline in the form of Range-Angle (RA), Range-Doppler (RD) and Angle-Doppler (AD) maps. These maps are sliced views of the total 3D Range-Angle-Doppler (RAD) cube, and obtaining any two combinations allows for the calculation of the third.
Unlike LiDARs, frequency-modulated continuous wave radars operate in the millimeter wave band in which signals do not get significantly attenuated when faced with occlusions, allowing operation ranges of up to 3,000m.
Radars function in adverse weather conditions more robustly than other commonly used sensing methods like cameras and LiDARs. Radar signals are also rich in information as they contain Doppler information that includes the velocity of the objects.
These radar features have motivated its usage not only in deterministic instrumentation but also for computer vision tasks \cite{rodnet, raddet}.
The radar signals can be processed to be used in an image-like pipeline in the form of Range-Angle (RA), Range-Doppler (RD), and Angle-Doppler (AD) maps. These maps are sliced views of the total 3D Range-Angle-Doppler (RAD) cube, and obtaining any two combinations allows for the calculation of the third.

% As stated earlier, LiDARs are the more popular approach for remote sensing used in automotive perception, both in the tasks of object detection and semantic segmentation. That said, t
The task of semantic segmentation using raw/processed radar data has been a growing task in the radar perception community and has shown promising development in recent years \cite{TRODNet, rodnet, raddet, radar_scenes, carrada, intro1, 9299052, 8904734, 9048939}. 
%The growing need for lower-cost, fast, and error-resistant sensing instruments mixed with compact and efficient deep-learning models is the main driver of this work.
Nonetheless, segmenting radar images still poses a challenge due to the noisy and sparse nature of the data, as well as the high imbalance between the foreground and background.
Also, despite the information-rich nature of radar data and the ability to obtain multiple views from a single sensing instance, most works do not utilize these benefits and tend to limit their approaches to Convolutional Neural Network (CNN) models on a single view, resulting in models that do not adequately capture global information from these maps. 
% Previous works in the literature explored the use of radar maps in semantic segmentation tasks, however, they have been limited in their approaches to neural networks or CNN methods.
To circumvent that, we propose a novel attention-based approach for semantic segmentation using radar data signals in radar learning. Our technique extends the definition of attention models to apply attention to adaptively sampled variations of our input feature maps, tackling the sparse nature of radar data. 
The adaptability nature of our attention block allows it to attend to multiple views of the Range-Angle-Doppler (RAD) cube in an efficient way.
 % To the best of our knowledge, there has not been a work proposing a paradigm similar to our method. 
 We also combine our model with a loss function tailored to sparse and highly imbalanced data of our task. We propose a combination of class-agnostic, multi-class, and multi-view consistency losses. \\
%general idea, work being done, motivation-isc, and summary
%this part after overall arch and limitations and motivation
\textbf{Contribution:} In this work, we propose an automotive radar sensing method that outperforms previous state-of-the-art works and sets new top scores in the reported metrics (Figure \ref{fig:intro_fig}). Our main contributions are:
\begin{itemize}
 \setlength\itemsep{0.1cm}
    \item We introduce a novel adaptive-directional attention block that efficiently captures information from a sparse receptive field and simultaneously tackles the multi-input multi-output nature of our task.
    \item We propose a novel loss function for the radar semantic segmentation task tailored to address the inherent main drawbacks of radar data. These drawbacks include the noisy and sparse nature of radar signals and the disproportional level of background/foreground objects. 
%Exploring the effect of the loss functions on learning and radar perception, and proposing a loss function tailored to address and mitigate the drawbacks of radar data.  
    \item Our proposed approach results in state-of-the-art performance in radar semantic segmentation of two recent datasets for radar perception, CARRADA \cite{carrada} and RADIal \cite{radial}, and achieves state-of-the-art results in the object detection task of the RADIal dataset.
\end{itemize}

%split into paragraph
\section{Related Work}
Low-cost frequency modulated continuous wave radars have been historically used in multiple applications involving machine learning and pattern recognition such as human activity and hand gesture recognition \cite{hand-gesture2,hand-gesture3,hand-gesture4}. In the context of automotive driving and autonomous vehicles, LiDAR sensors are more popular %and their data output usually comes in the form of point-cloud data. 
with a common data output in the form of a point cloud.
% While radar signals are rich in information, the way they are processed yields a different physical representation than what LiDARs provide. Nonetheless, multiple works explored point-cloud fusion of radars and LiDARs \cite{9000872,radsegnet}. 
While multiple works have explored point-cloud fusion of radars and LiDARs \cite{9000872,radsegnet}, radar signals processing usually yields different physical representation than the LiDAR.

The low resolution and high sparsity of radar data make the point-cloud format and associated architectures unsuitable. While some datasets provide point-cloud radar data \cite{radar_scenes,robotcar}, recent approaches to radar processing use the full/split processed RAD tensors in the shape of 3D/2D image-like data. Common radar datasets provide either a single view of the data (either RA or RD) \cite{rodnet,radial}, the original raw and unprocessed radar signals \cite{radial}, or the full RAD tensors \cite{raddet, 2021multiview}. RAD tensors provide cohesive information of the radar data; however, it is often undesirable to use 3D data due to the increased complexity of models when associated with the density of radar data, especially when taking multiple frames from the temporal domain. In this work, we focus our efforts on getting an automated radar perception model through sliced radar RAD tensors and comparing our method to similar works.

% Considering the recency of radar semantic segmentation, there are a few competing methods that set a baseline for our approach. Detailed comparisons of performances and parameters will be provided in Section \ref{quantitative_results}. 
With the recent emergence of radar datasets \cite{carrada,radial}, few methods have been proposed for semantic segmentation and object detection.
% Due to the lack of research in this field, most of the competing baselines, like UNet and DeepLab3+, were not created for radar perception and they were fit as a backbone to set a baseline for the discussed methods. 
While common methods for image semantic segmentation can be employed, such as UNet \cite{unet }and DeepLabv3+ \cite{deeplabv3+}, these methods are not tailored to the noisy and sparse nature of radar images.
% We highlight the most recent and relevant works , which are the TMVANet \& MVNet \cite{2021multiview}, RAMP-CNN \cite{rampcnn}, T-RODNet \cite{TRODNet}, and PeakConv \cite{zhang2023peakconv}. 
We highlight the most recent and relevant works that process radar data. 
TMVA-Net \cite{2021multiview} is a multi-view method that is composed of an encoding block, a latent-space processing, and a decoding block. 
% It fully consists of convolutional layers and set a baseline of predictions in RD and RA maps near the release of the dataset. 
It fully consists of convolutional layers and presents a strong baseline for predictions in RD and RA maps on the CARRADA dataset.
RAMP-CNN \cite{rampcnn} is a CNN-based model that was mainly designed for processing 3D RAD tensors but was re-purposed for this dataset. 
T-RODNet \cite{TRODNet} is a recent model utilizing Swin Transformers \cite{liu2021Swin} but does not produce RD predictions and operates only on RA inputs. 
% hence, we will focus on TMVANet as the most recent competing SOTA model while citing T-RODNet as SOTA for RA scores.
While T-RODNet shows improved RA scores, we focus on simultaneous prediction of the RD and RA semantic segmentation maps.
PeakConv \cite{zhang2023peakconv} applies the convolution operation with a receptive field consisting of the peaks of the signal. While this approach achieves improved segmentation performance compared to TMVA-Net, it also increases the number of parameters.

Sparse variants of attention have been proposed in the literature. ReLA \cite{xu2022sparse} replaces the softmax activation with ReLu to achieve sparsity in attention and uses layer normalization to improve translation tasks. The sparsity can range from switching off attention to applying attention to all the input. On the other hand, our method learns the offsets to which the attention is applied and targets consistent efficiency for the radar segmentation task. Other sparse attention methods, such as NPA \cite{xue2022efficient} and SCAN \cite{zhang2021sparse} address point clouds, which are sparse in nature. Our method aims at learning to select important locations in the radar map dense grid.

\begin{figure*}
\begin{center}

\includegraphics[width=.95\linewidth]{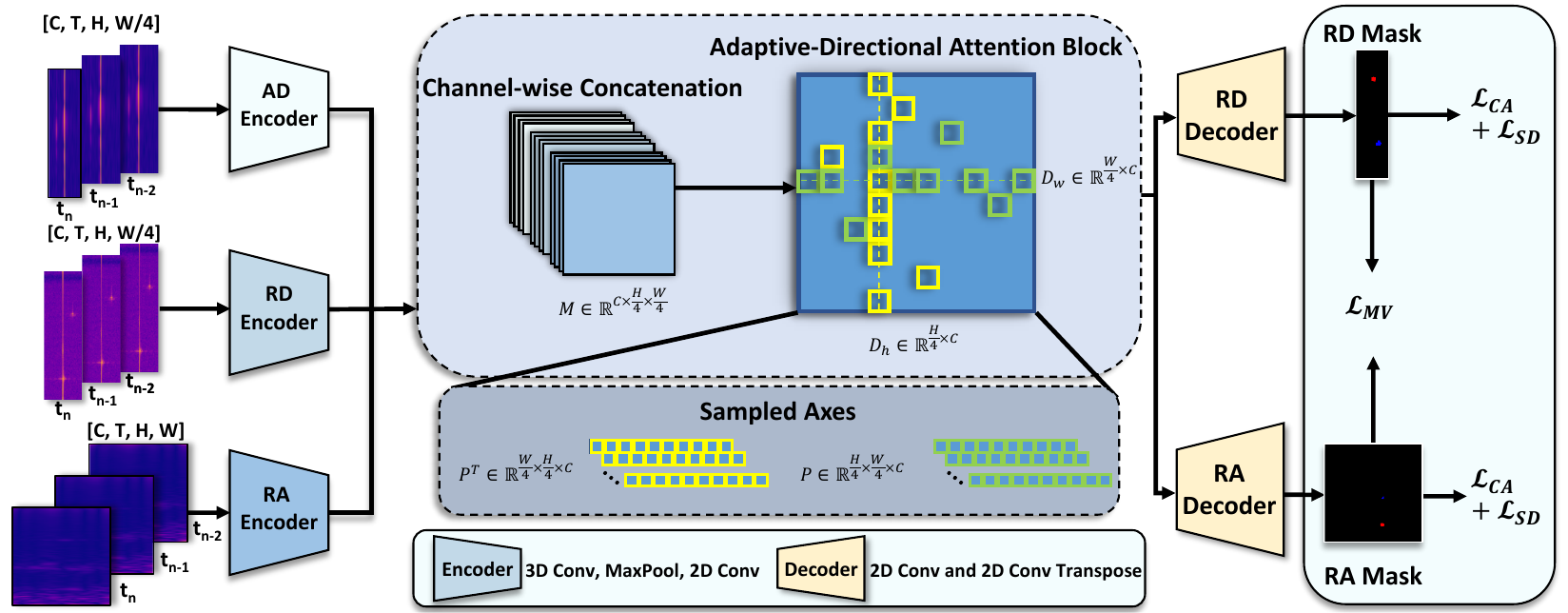}
\end{center}
\vspace{-4mm}
   \caption{Overview of our proposed method for radar semantic segmentation. The model starts by encoding multiple frames of the Angle-Doppler (AD), Range-Doppler (RD), and Range-Angle (RA) maps. The encoded features are concatenated into a single block of feature maps that is then passed into our adaptive-directional attention block. The adaptive-directional attention blocks sample rows and columns following Eqs. \ref{eq:sampling1} \& \ref{eq:sampling2} and apply self attention following Eq. \ref{eq:self_attention} after each sampling instance. The outputs are then split into two decoders generating RD and RA masks that are passed into our three loss functions described in Section \ref{total_loss}.}
\label{fig:full_model}
\end{figure*}

\section{Baseline} \label{baseline}% second 
% Despite T-RODNet \cite{TRODNet} reporting the SOTA results for RA predictions, the model does not produce RD mask predictions. Therefore we will focus on TMVANet as the most recent competing SOTA model.

TMVA-Net starts by encoding the RA, RD, and AD input maps to reduce the input size to one-fourth of its original resolution.
%from $1 \times T \times 256 \times 64$ for the AD and RD maps, and $1 \times T \times 256 \times 256$ for the RA map, to a common dimension of $128 \times 64 \times 64$.%
Each output is then passed into an Atrous Spatial Pyramid Pooling (ASPP) block \cite{aspp}, and is also concatenated into a single feature maps holder. Both the ASPP output and the concatenation are then passed into a two-branches (RA and RD) decoding space that produces prediction maps. TMVA-Net uses a combination of three loss functions: a weighted Cross-Entropy loss, where the weights correspond to the frequency of classes in the dataset, a weighted Soft Dice loss, and a coherence loss. The coherence loss is a mean-square error of the RD and RA outputs to ensure coherence of predictions from different views.
% %remove
% The second loss is the Soft Dice loss (SD). SD loss is commonly used in semantic segmentation tasks \cite{softdice, dice} and yields good results, making it a good fit in radar semantic segmentation. The SD loss is defined as:\

% \begin{equation}

% This results in our localization loss (LL) being equal to:

%     \mathcal{L}_{SD} = \frac{1}{K} \sum^{K}_{k=1}[1- \frac{2\sum \textbf{y}\textbf{p}}{\sum \textbf{y}^2 + \textbf{p}^2}]
% \end{equation}
% where $\textbf{. The difference between this term and the previous one is that: 1) this is used on multi-class, so we heavily penalizey}$ and $\textbf{p}$ refer to the ground truth and probability map output of the model, respectively. 

% The last loss that TMVANet incorporates is a coherence loss, whose purpose is to make sure that the Range (R) information in both RA and RD map outputs are coherent. The rationale here lies in preserving the consistency between physically defined values. The coherence loss is defined as:

% \begin{equation}
%     \mathcal{L}_{CoL}(\textbf{p}^{RD}, \textbf{p}^{RA}) =  ||\textbf{p}^{\sim RA} - \textbf{p}^{\sim RA}||^2_F ,
% \end{equation}
% ($\sim$) refers to the prediction map after eliminating (max-pooling) the second dimension of both tensor, leaving the range (R) dimension behind.

\subsection{Limitations} %focus on its limitations to create the motivation

The mentioned models yield state-of-the-art results in radar semantic segmentation on the CARRADA dataset. Nonetheless, these models have limitations pertaining to the nature of the implementation and the task. 
%First, the multi-input structure of the dataset results in the model learning only spatial information propagating through the data if the models are limited to using convolutions. 
First, the models are limited to convolution layers that learn local spatial information of the multi-input data.
% Furthermore, to increase the accuracy of said models, increasing the number of feature maps was necessary at every layer of the model, up to a certain point where the model reaches its maximum capacity.
While increasing the number of feature maps at every layer would slightly improve the accuracy of these models, it imposes a large computation burden. 
% Attempting to reduce the feature maps results in a proportionally lower mIoU score. 
This impedes the model from further improving without increasing the number of parameters with the majority of parameters being employed in the convolutional layers. % with large feature maps. 
The second limitation is the ability of these models to learn and retain information from other maps. T-RODNet processes RA maps only, while TMVA-Net concatenates all feature maps in the bottleneck along with the ASPP outputs. For the rest of the model, all combined feature maps are treated as a single set of feature maps coming from one source that gets split into two prediction heads. 
% Another limiting aspect of said models is the disparity in the number of parameters. 

Another important aspect to be considered in these methods is the number of parameters.
TMVA-Net produces multi-view results with $50\times$ less parameters than T-RODNet. Lastly, all reported models were trained using the combination of losses which are not optimally designed for the task of radar semantic segmentation. Therefore, we propose an alternative approach in Section \ref{total_loss}.
% The second limitation is the ability of these models to learn and retain information from other maps. T-RODNet looks at RA maps only, while TMVANet concatenates all feature maps in the bottleneck along with the ASPP outputs. For the rest of the model, all combined feature maps are treated as a single set of feature maps coming from one source that gets split into two prediction heads. Another limiting aspect of said models is the disparity in the number of parameters. TMVANet produces SOTA results with only $5.6\times10^6$ parameters, while other models like T-RODNet and RAMP-CNN are 50$\times$ larger than TMVANet. We note that while T-RODNet proposes an attention-based model, its significantly larger model size makes it an undesirable approach. All reported models were trained using the combination of losses mentioned in Section \ref{baseline}, which are not optimally designed for the task of radar semantic segmentation and we propose an alternative approach in Section \ref{total_loss}.
%%%%
\vspace{-0.1cm}
\section{The Proposed Method}
\subsection{Motivation} %third

Our proposed method is designed to address the limitations we observed in the state-of-the-art models discussed previously. We aim to create a compact model that improves upon previous methods by addressing the issues observed in model learning through a proposed novel architecture and loss functions. Our method overcomes the hurdle of introducing attention in deep learning models by minimizing the number of tokens to keep the model fast and small. We also take into consideration the sparse nature of the radar data while implementing our method. We propose a loss function tailored specifically for the task of radar learning by taking the acquisition structure into consideration. We extend our approach to addressing the issue of class imbalance in a more refined way compared to weighted cross-entropy, and we tackle the poor localization ability of the proposed models in our loss functions. Lastly, we propose a new multi-view range matching loss that addresses the drawbacks of fused multi-view inputs.

\subsection{Overall Architecture} % first with figure and caption
We propose a lightweight attention-based neural network architecture, shown in Figure \ref{fig:full_model}, which addresses the limitations of the previous works. The model starts by using a similar encoding module as the one used in TMVA-Net \cite{2021multiview}, with $x_i \in \mathbb{R}^{1\times T\times H\times W}$ where $x_i$ is an RA, RD, or AD feature map, $T$ is the number of past frames taken from the range $[t_0 - T, t_0]$, and $H$ and $W$ are the height and width of the radar frequency map, respectively. The feature maps generated from the encoders are expressed as $x_{en} \in \mathbb{R}^{C\times H_d\times H_d}$, where $x_{en}$ is an encoded feature map, $C$ is the number of feature maps, and $H_d$ and $W_d$ are the downsampled heights and widths, respectively. The produced feature maps are then channel-wise concatenated into a single latent space that constitutes the input to our adaptive-directional attention block. 
%In convolution-based competing methods, reducing the feature maps below $128$ channels in the latent bottleneck reduces the mIoU considerably, while our attention-based approach requires less than half that number to produce similar scores. 
In convolution-based competing methods, we noticed that reducing the feature maps below $128$ channels in the latent bottleneck greatly reduces the mIoU, so we adopt an attention-based approach that achieves similar scores with smaller feature maps.

Contrary to other attention-based approaches in radar perception \cite{TRODNet}, we do not need to use convolutional layers or heavy positional embeddings. Instead, we shed light on the way the dataset is constructed, where the multi-view input has implicit information that can be shared across axes and channels. Figure \ref{fig:full_model} illustrates the operation mechanism of our adaptive-directional attention block after the concatenation of the inputs' encoding.

% The concatenated output of the encoding goes through a 2D deformable convolution. Deformable convolutions have proven to be more potent in establishing spatial connections between the multi-view axes before the attention when compared to normal convolutions. The output of the deformable convolution layer gets permuted into a $W\times H \times C$ tensor, where the multi-headed attention is then applied on the flattened embedding of $W\times C$ tensor. The output gets reshaped back into a $C\times H\times W$, followed by another deformable convolution. The second attention layer follows the same procedure, but the tensor is permuted to $H\times W \times C$. 

\subsection{Adaptive-Directional Attention}

In our model architecture, we propose a novel adaptive-directional attention block that composes the backbone of our model. Similar concepts of sampling straight-vector axes were previously proposed in the literature \cite{ccnet,axialdeeplab,axial}. % However, our implementation focuses on the adaptive sampling of significant variations of the axes and applying the attention on a sample-by-sample basis. 
However, our adaptive-directional attention tackles the sparse nature of radar data by utilizing attention that can extend further than single-column/row attention. In this way, it ensures a comprehensive outlook of the information space while being computationally efficient.
For a 2D input image of shape $C \times H_{d} \times W_{d}$, we obtain two attention variations, one of the shape $H_{d} \times W_{d} \times C$ and another of the shape $W_{d} \times H_{d} \times C$. For example, for a width $W_{d}$, we have $W_{d}$ sampled vectors of size $H_{d} \times C$. 
% The rationale behind incorporating the channels in our sampling traces back to our explanation regarding radar data's feature maps being rich in information. 
The rationale behind incorporating the channels in our sampling traces back to the rich information provided by the radar data's feature maps. 
%Our adaptive-directional attention was inspired by previous works in deformable attention transformers such as Deformable DETR \cite{detr} and Deformable Attention Transformers \cite{deformable_attention}.
We sample our axes by employing vertical and horizontal iteration limits of sizes $k_h$ and $k_w$, respectively. We also define the horizontal and vertical shifts, $\Delta h$ and $\Delta w$, that constitute the offset limits of sampling. Lastly, we define learnable parameters $\theta_h$ and $\theta_w$ that perform a modulating operation to limit the effect of noise seen in data, allowing the model to learn to suppress insignificant regions. 
% We noticed that introducing a third dimension to the adaptive sampling and including channel modulating considerably increases the model complexity without a significant increase in performance. 
Using these definitions, we then write the sampling operation that occurs before the attention on the columns as:

\begin{equation}\label{eq:sampling1}
    x_{i,j} =  \sum^{w}_{k=1} {(\theta_h)_k \cdot X_{H,C}^{(i,j+\Delta h_k)}} 
\end{equation}
% \begin{equation}
%     x_{i,j} =  \sum^{w}_{k=1} {(\theta_h)_k \cdot X_{H,C}^{(i+\Delta h_k,j)}} 
% \end{equation}
where $x_{i,j}$ is the value of the column with indices $i,j$ belonging to the axes as $i \in [0,H]$ and $j \in [0,C]$. Parameter $w$ refers to the horizontal iterations limit (i.e. how many pixels we iterate over), belonging to the previously defined parameter $k_w$. $(\theta_h)_w$ is the corresponding modulation weight for the associated shift, and $\Delta h_w$ covers how far we sample from the axis center (i.e. the starting column). 

After the sampling operation, we obtain $W_{d}$ vectors of size $H_{d} \times C$. The query, key, and values (\textbf{q}, \textbf{k}, \textbf{v}) are then obtained through multi-layer perceptron layers, where the multi-headed self-attention (MSA) is then calculated as:

\begin{equation}\label{eq:self_attention}
\begin{split}
    &SA(q,k,v) = Softmax(\frac{qk^T}{\sqrt{d_k}})v \\
    &MSA = [SA_1; SA_2; ...; SA_s] 
\end{split}
\end{equation}
for $s$ heads obtained from the input, following the formulation in vision transformers \cite{dosovitskiy2020vit}. We note that we first sample by columns (i.e. produce $W_{d}$ vectors of size $H_{d} \times C$) and apply MSA, then sample by rows (i.e. produce $H_{d}$ vectors of size $W_{d} \times C$) and apply the second MSA. The formulation for the MSA applied to the rows is similar to that of the columns, with the following row sampling:%, where row sampling is done using a similar logic to the columns, shown as:

\begin{equation}\label{eq:sampling2}
    x_{i,j} =  \sum^{h}_{k=1} {(\theta_w)_k \cdot X_{W,C}^{(i+\Delta w_k,j)}} 
\end{equation}

Unlike convolution-based transformers or other types of attention modules, the nature of our adaptive-directional attention allows us to alleviate the need for convolutional channel mixing or expansions. The adaptive sampling reduces the model complexity significantly by incorporating a convolution-like operation before applying attention.

\subsection{Proposed Loss Function} \label{total_loss}%class agnostic object-background loss 

Model learning in both semantic segmentation and object detection can prove difficult due to the large ratio of background to foreground pixels. This disparity was historically studied in multiple works that addressed the issue either through employing multi-stage detectors \cite{feature_pyramid_object_detector, region_based_object_detector} in object detection, or targeting the way models learn through innovative loss functions that handle class imbalance in semantic segmentation \cite{focal_loss_object_detector,unified_focal_loss}. Radar-based datasets have a larger proportion of background pixels when compared to actual objects (foreground). This discrepancy is notably present in the datasets we operate on, where the background class consists of more than 99\% of the total dataset pixels \cite{carrada, radial}. In addition to the class imbalance between background and foreground pixels, the annotated objects are relatively small in pixel size. Lastly, RD, RA, and AD maps' noisy nature is a learning hurdle for the models. To tackle these issues, we propose an Object Centric-Focal loss (OC) and a Class-Agnostic Object Localization Loss (CL). We add both of them in a single term, the Class-Agnostic Object Loss (CA), and propose a new multi-view range matching loss (MV) that suits our multi-output architecture.

%\cite{focal_loss_object_detector}\cite{tversky}

\subsubsection{Class-Agnostic Object loss}

\textbf{Object Centric-Focal Loss}: The main highlight of this loss is the weighing of the binary cross-entropy between the background and foreground of the predictions, with higher weight being given to the foreground. This is defined as:

\begin{equation}\label{eq:OC}
\mathcal{L}_{OC} = (1 - y_{pred}) (\delta \mathcal{L}_{BCE_{FG}} + (1- \delta)\mathcal{L}_{BCE_{BG}})
\end{equation}
where $\delta$ is a weighing factor (set to 0.6) and $\mathcal{L}_{BCE}$ is the binary cross entropy, calculated with the two classes 'background' and 'foreground'. While our semantic segmentation objective includes multi-class labels, we aim to use this loss to penalize the model on hard background prediction, keeping it only to a binary background/foreground calculation. While other loss functions \cite{focal_loss_object_detector} propose a power factor on the $(1 - y_{pred})$ term, we instead remove it and use one-hot prediction masks. Both operations come in favor of having a balanced approach between ground truth probabilities and loss value, and heavily penalizing misclassification between the background and foreground.

\begin{figure}
\begin{center}

\includegraphics[width=1\linewidth]{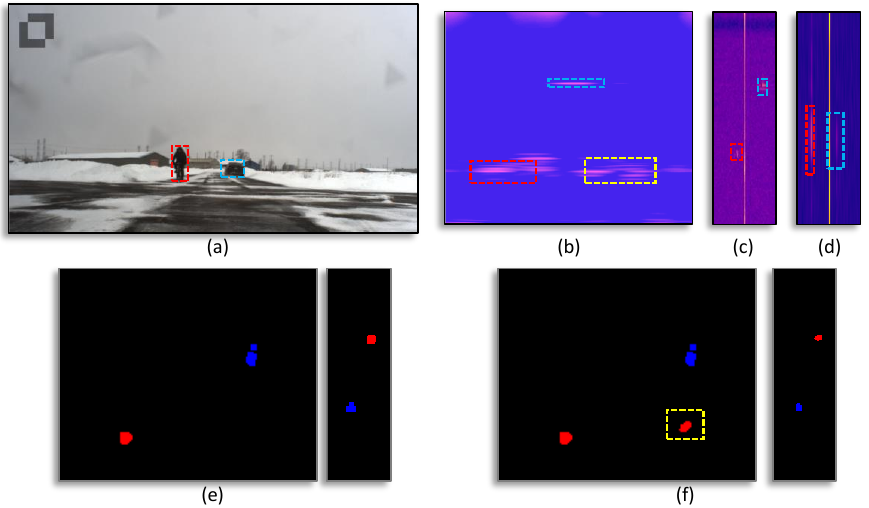}
\end{center}
\vspace{-4mm}
   \caption{Radar (b) RA, (c) RD, and (d) AD maps with (a) synchronized RGB image. Red and blue annotation boxes correspond to the person and car, respectively, shown in the RGB image. We highlight a sample random noise appearing on the RA map with a yellow box. (e) shows the ground truth mask for the RA and RD maps (left to right) of this scene, and (f) shows a false segmentation with the noise seen as an object. The noise shown in the RA map does not appear as frequently in RD maps. The contrast of the maps was edited for illustration purposes.}
\label{fig:noise_presence}
\end{figure}

\noindent\textbf{Class-Agnostic Object Localization Loss}: To illustrate the rationale of proposing this localization loss, we show RA and RD input maps with their output predictions, along with the corresponding RGB image in Figure \ref{fig:noise_presence}. Any other object signature seen in the RA input image can be attributed to speckle noise, Doppler-induced noise, or any other sort of undesired noise that is unaccounted for. Due to this noisy nature of radar data, producing a significantly larger amount of false positives was a noticeable pattern across tested models. We also noticed similar behavior in the opposite way, where the model learns the noise as part of the background and confuses objects with similar signatures as the noise for being part of the background, resulting in many false negatives. 
%This results in us needing an intersection-based loss that penalizes the model on false background/foreground predictions. 
Therefore, we propose an intersection-based loss that penalizes the model on false background/foreground predictions.
This builds on the previous object-centric loss by creating an IoU-based loss that penalizes mislocalization of objects, defined as:

\begin{equation}\label{eq:CL}
    \mathcal{L}_{CL} = 1 - \frac{TP}{TP + FN + FP},
\end{equation}
where $TP$ refers to true positives, $FN$ to false negatives, and $FP$ to false positives. Similar to $\mathcal{L}_{OC}$, we extend our implementation to focus on the one-hot predictions instead of the probability maps, %which is harsher when penalizing the model for making a faulty background prediction. 
which imposes a larger penalty for making a false background prediction. 
Adding $\mathcal{L}_{OC}$ and $\mathcal{L}_{CL}$ terms yields our class-agnostic object loss: $\mathcal{L}_{CA} =  \mathcal{L}_{OC} + \mathcal{L}_{CL}$.

% \begin{equation}\label{eq:CA}
%     \mathcal{L}_{CA} =  \mathcal{L}_{OC} + \mathcal{L}_{CL},
% \end{equation}

\subsubsection{Multi-Class Segmentation  Loss}

 To include the multi-class nature of our dataset and localization of different class predictions, we use a similar Soft Dice loss (SD) term to the one used in \cite{2021multiview}, described as:

\begin{equation}
    \mathcal{L}_{SD} = \frac{1}{K} \sum^{K}_{k=1}[1- \frac{2\sum \textbf{y}\textbf{p}}{\sum \textbf{y}^2 + \textbf{p}^2}]
\end{equation}
where $\textbf{y}$ and $\textbf{p}$ refer to the ground truth and probability map output of the model. Unlike the previous terms, we do not use a one-hot binary map prediction and instead use the original continuous probability map. We also do not limit $\mathcal{L}_{SD}$ to background/foreground classes since we use it for multi-class predictions.

% This loss term reduces both FN and FP to make the loss significant, with FN being higher for values of $\delta$ larger than $0.5$, which is the range we operate at. To illustrate the rationale of proposing this loss, we show an RA map sample synchronized to an RGB image in Figure \ref{fig:noise_presence}. Any other object signature seen in the image can be attributed to speckle noise, Doppler noise, or any other sort of undesired noise that is unaccounted for. Due to this noisy nature of radar data, producing a significantly larger amount of false positives is a noticeable pattern across tested models. We also noticed similar behavior in the opposite way, where the model learns the noise as part of the background and confuses objects with similar signatures as noise for being part of the background, resulting in many false negatives. This results in $\delta$ being an important weighing factor that tweaks the loss to minimize this noisy behavior. Similar to our CAML, we extend our implementation to focus on the one-hot predictions instead of the probability maps, which is less forgiving when penalizing the model for making a faulty background prediction.

% We then define CA as the sum of CAIL and CAML, where:

% \begin{equation}
%     \mathcal{L}_{CA} =  \mathcal{L}_{CAIL} + \mathcal{L}_{CAML},
% \end{equation}

% We explore in Section \ref{ablation} the effects of ignoring the CAIL term (i.e. CA is just CAML) and ignoring the CAML (i.e. CA is just CAIL) on the model scores.

\subsubsection{Range Consistency Loss}
In addition to the class-agnostic object loss and multi-class segmentation loss, we define a Multi-View range matching loss (MV) as: 

\begin{equation}
\mathcal{L}_{MV}=
    \left\{\begin{matrix}
        \frac{1}{2}(RD_m - RA_m)^{2} & \left |RD_m - RA_m \right | < 1\\
        |RD_m - RA_m| - \frac1 2 & otherwise
    \end{matrix}\right.
\end{equation}
where $RD_m$ and $RA_m$ are the max-pooled RA and RD probability maps, leaving only the $R$ direction. The analytical term of this loss is a special case of the Huber loss \cite{huber} and was proven to be more robust than mean-square error when dealing with outliers. \\[1mm]%This improved RA score without reducing RD's as will be shown in Section \ref{ablation}.\\[1mm]
\noindent\textbf{Overall Loss:} Our total loss is then defined as the weighted sum of all proposed losses with weights $\alpha_1$, $\alpha_2$, and $\alpha_3$ as:
\begin{equation}
    \mathcal{L}_{total} = \alpha_1 \mathcal{L}_{CA} + \alpha_2 \mathcal{L}_{SD} + \alpha_3 \mathcal{L}_{MV}
\end{equation}
% \begin{equation}
%     \mathcal{L}_{coh}(\textbf{p}^{RD}, \textbf{p}^{RA}) =  \mathcal{L}_{H}(\textbf{p}^{RA},\textbf{p}^{RA}) ,
% \end{equation}

% \subsection{Training and Inference}

\begin{table*}[h]
\footnotesize
\centering
\begin{tabular}{l l l  c c c c |c|  c c c c |c}
\hline
\multirow{2}{*}{View} & \multirow{2}{*}{Method} & \multirow{2}{*}{Params (M)} & \multicolumn{5}{c}{IoU (\%)} & \multicolumn{5}{c}{Dice (\%)} \\
\cmidrule(lr){4-8}\cmidrule(lr){9-13}
& & & \multicolumn{1}{c}{Bkg.} & \multicolumn{1}{c}{Ped.} & \multicolumn{1}{c}{Cycl.} & \multicolumn{1}{c}{Car} & \multicolumn{1}{c}{mIoU} & \multicolumn{1}{c}{Bkg.} & \multicolumn{1}{c}{Ped.} & \multicolumn{1}{c}{Cycl.} & \multicolumn{1}{c}{Car} & \multicolumn{1}{c}{mDice}\\
\hline  
\multirow{8}{*}{RD} & FCN-8s \cite{fcn8} & 134.3 & 99.7 & 47.7 & 18.7 & 52.9 & 54.7 & 99.8 & 24.8 & 16.5 & 26.9 & 66.3 \\
& U-Net \cite{unet} & 17.3 & 99.7 & 51.1 & 33.4 & 37.7 & 55.4 & 99.8 & 67.5 & 50.0 & 54.7 & 68.0\\
& DeepLabv3+ \cite{deeplabv3+} & 59.3 & 99.7 & 43.2 & 11.2 & 49.2 & 50.8 & 99.9 & 60.3 & 20.2 & 66.0 & 61.6 \\ 
   & RSS-Net \cite{wce} & 10.1 & 99.3 & 0.1 & 4.1 & 25.0 & 32.1 & 99.7 & 0.2 & 7.9 & 40.0 & 36.9 \\ 
    & RAMP-CNN \cite{rampcnn} & 106.4 & 99.7 & 48.8 & 23.2 & 54.7 & 56.6 & 99.9 & 65.6 & 37.7 & 70.8 & 68.5 \\ 
     & MVNet \cite{2021multiview} & 2.4 & 98.0 & 0.0 & 3.8 & 14.1 & 29.0 & 99.0 & 0.0 & 7.3 & 24.8 & 32.8 \\ 
    & TMVA-Net \cite{2021multiview} & 5.6 & 99.7 & 52.6 & 29.0 & 53.4 & 58.7 & 99.8 & 68.9 & 45.0 & 69.6 & 70.9 \\ 
   & PeakConv \cite{zhang2023peakconv} & 6.3 & - & - & - & - & 60.7 & - & - &  & - & 72.5 \\
 \cmidrule{4-13}
& \textbf{TransRadar} & 4.8 & \textbf{99.9} & \textbf{57.7} & \textbf{36.1} & \textbf{61.9} & \textbf{63.9} & \textbf{99.9} & \textbf{73.2} & \textbf{53.1} & \textbf{76.5} & \textbf{75.6} \\[0.1ex] 
\hline
\multirow{9}{*}{RA} & FCN-8s \cite{fcn8} & 134.3 & 99.8 & 14.8 & 0.0 & 23.3 & 34.5 & 99.9 & 25.8 & 0.0 & 37.8 & 40.9 \\
& U-Net \cite{unet} & 17.3 & 99.8 & 22.4 & 8.8 & 0.0 & 32.8 & 99.9 & 25.8 & 0.0 & 37.8 & 40.9\\
& DeepLabv3+ \cite{deeplabv3+} & 59.3 & 99.9 & 3.4 & 5.9 & 21.8 & 32.7 & 99.9 & 6.5 & 11.1 & 35.7 & 38.3 \\ 
   & RSS-Net \cite{wce} & 10.1 & 99.5 & 7.3 & 5.6 & 15.8 & 32.1 & 99.8 & 13.7 & 10.5 & 27.4 & 37.8 \\ 
    & RAMP-CNN \cite{rampcnn} & 106.4 & 99.8 & 1.7 & 2.6 & 7.2 & 27.9 & 99.9 & 3.4 & 5.1 & 13.5 & 30.5 \\ 
     & MVNet \cite{2021multiview} & 2.4 & 98.8 & 0.1 & 1.1 & 6.2 & 26.8 & 99.0 & 0.0 & 7.3 & 24.8 & 28.5 \\ 
    & TMVA-Net \cite{2021multiview} & 5.6 & 99.8 & 26.0 & 8.6 & 30.7 & 41.3 & 99.9 & 41.3 & 15.9 & 47.0 & 51.0 \\ 
    & T-RODNet \cite{TRODNet} & 162.0 & 99.9 & 25.4 & 9.5 & \textbf{39.4} & 43.5 & 99.9 & 40.5 & 17.4 & \textbf{56.6} & 53.6 \\ 
    & PeakConv \cite{zhang2023peakconv} & 6.3 & - & - & - & - & 42.9 & - & - &  & - & 53.3 \\
 \cmidrule{4-13}
& \textbf{TransRadar} & 4.8 & \textbf{99.9} & \textbf{30.3} & \textbf{21.5} & 38.2 & \textbf{47.5} & \textbf{99.9} & \textbf{46.6} & \textbf{35.3} & 55.3 & \textbf{59.3} \\[0.1ex] 
    \hline
\end{tabular}
\caption{Semantic segmentation performance on the test split of the CARRADA dataset, shown for the RD (Range-Doppler) and RA (Range-Angle) views. Columns from left to right are the view (RD/RA), the name of the model, the number of parameters in millions, the intersection-over-union (IoU) score of the four different classes with their mean, and the Dice score for the same classes.}
\label{tab:full_results}
\end{table*}

\section{Experiments} 
\subsection{Datasets}
To test the effectiveness of our proposed approach, we use the CARRADA \cite{carrada} dataset as the main multi-class radar semantic segmentation dataset. 
%We also use RADIal \cite{radial} dataset to further assert our model's ability to achieve SOTA results in radar semantic segmentation.
We also test our proposed method on RADIal \cite{radial} dataset and compare to previous state-of-the-art methods in radar semantic segmentation and object detection.

\textbf{CARRADA:} The CARRADA \cite{carrada} dataset consists of synchronized camera-radar recordings of various driving scenarios containing 12,666 frames. The annotations of the data were done semi-automatically and provided for the RD and RA views \cite{carrada}. The dataset contains four object categories: pedestrian, cyclist, car, and background. The input are the RA, RD, and AD maps decomposed from the 3D RAD tensor. RA maps have a size of $1\times 256\times 256$ while RD and AD have a different resolution of $1\times 256 \times 64$. We use the 2D decomposition of the RAD tensor to reduce the model complexity, which is an important factor in radar perception in automotive driving.

\textbf{RADIal:} The RADIal \cite{radial} dataset is a new high-resolution dataset consisting of 8,252 labeled frames. RADIal varies from CARRADA in that it does not provide a multi-view input and depends only on RD input. The outputs are also produced and compared to projected annotated RGB images, unlike the CARRADA dataset that compares annotation directly in the RD/RA planes. RADIal also provides a high-definition input, where the input size is $32\times512\times256$.
% where the channels represent $N_{R_{X}} = 16$  antenna receivers, each receiving a complex number signal consisting of two number $(C_{MIMO} = 2 \times N_{R_{x}})$. 
RADIal provides annotations for two classes only: free-driving-space and vehicle annotations (i.e. free or occupied). 
\subsection{Evaluation Metrics}
We follow the same evaluation metrics used in previous works, which are the common intersection over union (IoU), the Dice score (F1 score), and the mean of each across the classes. The mIoU is also used to evaluate the semantic segmentation task on the RADIal dataset. The combination of the mIoU and the Dice score creates a fair and comprehensive assessment of the results. For the object detection task in RADIal, we use the same metrics as \cite{radial} with Average Precision (AP), Average Recall (AR), and regression errors.
% We follow the same evaluation metric used in previous works, which is the common intersection over union (IoU). For a given class $c$, the IoU is the percentage $\frac{|A_c\cap B_c|}{|A_c \cup B_c|}$ where $A_c$ is the set of predicted locations from class $c$ and $B_c$ is the set of ground-truth locations of class $c$. The average of this over all four classes yields the mean IoU (mIoU). This is also used as the evaluation metric of the semantic segmentation task in the RADIal dataset. We also use the same Dice score mentioned in \cite{2021multiview}, which is defined as $\frac{2|A_c\cap B_c|}{|A_c| + |B_c|}$ using the same explanation. The Dice score reports the F1 score, which complements the mIoU in creating a fair and comprehensive assessment of the results. While our model was not created for the object detection task, TransRadar still yielded SOTA results in RADIal object detection and localization which will be reported in Section \ref{quantitative_results}. The main metrics used in the object detection task are Average Precision (AP), Average Recall (AR), and regression errors, where minimizing the range and angle of object prediction is a positive outcome.

\subsection{Implementation Details}
We implement and train TransRadar using the PyTorch library on a single NVIDIA A100 GPU. All reported models on the CARRADA dataset were trained with a batch size of 6 and using 5 past frames. We use Adam optimizer \cite{adam}, initial learning rate of $1\times10^{-4}$, and an exponential scheduler (step = 10). For our final TransRadar model, we use $8\times$ cascaded blocks of our adaptive-directional attention block. For the testing, we use a batch size of 1 and a similar number of past frames. 

For the RADIal dataset training, %we alter our model to be a backbone to replace the one used in FFTRadNet \cite{radial}. 
we replace FFTRadNet \cite{radial} backbone with our proposed model.
We employ a single-view encoding/decoding paradigm similar to the one shown in Figure \ref{fig:full_model}. We use the same segmentation and detection heads from the FFTRadNet model, and the same optimizer and scheduling as CARRADA dataset training.

\subsection{State-of-the-art Comparisons}\label{quantitative_results}
%\subsubsection
\noindent\textbf{Semantic Segmentation  on the CARRADA:} \label{qualitative_results}
%\textbf{Quantitative Results:}
Table \ref{tab:full_results} shows the quantitative comparisons of the proposed approach with existing state-of-the-art frameworks for radar semantic segmentation. 
%, we report the same metrics and comparisons reported in previous state-of-the-art works in Table \ref{tab:full_results}. 
The results listed in the table show that TransRadar outperforms state-of-the-art methods in both the mIoU and mDice metrics. A large part of this is attributed to the introduction of the CA loss, which will be discussed in detail in the ablation studies in Section \ref{ablation}. Our model achieves new state-of-the-art performance with an RD mIoU score of 63.9\%, which outperforms the closest baseline by 3.2\%, and has a mDice score of 75.6\%. For the RA map predictions, our method yields a mIoU of 47.5\%, outperforming the state-of-the-art score by 4.0\%, with a mDice of 59.3\%. We also point out that our model significantly outperforms other models in the Cyclist class, where we note a large gap of 12.0\% between our model and the second-best model in the RA map, and 13.1\% in the RD map. This can be attributed to the consistency with RD as well as the ability to predict harder examples. Across the board, our model sets new state-of-the-art scores except for the car class IoU and Dice in the RA maps, where T-RODNet has a slightly higher score.

\begin{figure}
\begin{center}

\includegraphics[width=1\linewidth]{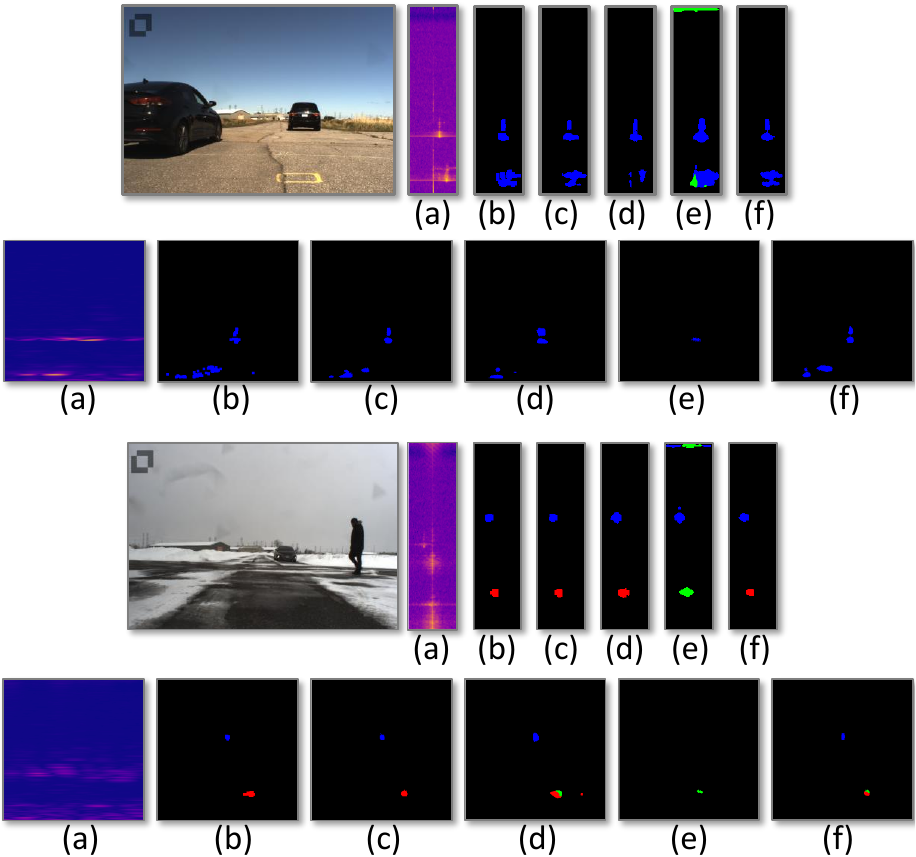}
      
\end{center}
\vspace{-4mm}
   \caption{Qualitative results on two test scenes from the CARRADA test split showing the RGB camera view with results of semantic segmentation from different methods. For every image, (top) depicts the RD and (bottom) depicts RA. (a) RD/RA inputs, (b) ground-truth, (c) TransRadar, (d) TMVA-Net \cite{2021multiview}, (e) MVNet \cite{2021multiview}, and (f) UNet \cite{unet}. All RD outputs were rotated for visual coherency. Different colors correspond to different classes. \textcolor{blue}{Blue: Car}, \textcolor{green}{Green: Cyclist}, \textcolor{red}{Red: Pedestrian}. Black: background.}
\label{fig:qualitative}
\end{figure}

%\textbf{Qualitative Results:}
Figure \ref{fig:qualitative} shows two qualitative results on a hard scene and a normal scene from the test split of CARRADA. The first scene shows a good segmentation with instances of mislocalization in all tested methods, with TransRadar and UNet giving the best prediction results. We then present a well-segmented RD and RA predictions in the second scene relative to the mask from our method when compared to other models. We also notice a coherent translation of the RD to RA views in the range dimension in both scenes.\\

%\subsubsection

\noindent\textbf{Semantic Segmentation on RADIal:}
%\subsection{RADIal}
%\textbf{Semantic Segmentation:}
We further look at the semantic segmentation results of the RADIal dataset shown in Table \ref{tab:radial_seg}. Our method outperforms all previously reported models in the semantic segmentation task with a mIoU of 81.1\% and less than half the model size of the most recently reported state-of-the-art method, C-M DNN \cite{radar_multitask_learning}. 
% This gives us confidence in claiming that our model is well tailored to tackle radar perception tasks.
These results showcase the ability of our proposed method, which is tailored to radar data, to tackle various datasets.\\
\noindent\textbf{Object detection on RADIal:}
%\textbf{Object Detection:}
Object detection results on the RADIal dataset are shown in Table \ref{tab:radial_od}. Our method outperforms all previously reported models in this task as well with significantly higher AR and lower angular prediction error. Despite our method not being designed for the task of object detection, the model still sets a new record for this task. All taken into account, our model sets a new standard for state-of-the-art predictions in these two datasets.

\begin{table}[h]  
\centering % centering table  
\begin{tabular}{l c c c c} % creating 10 columns  
\hline   
Backbone  & \# Params. (m) & \% mIoU  \\ [0.5 ex]  
\hline  
PolarNet \cite{polarnet} & - & 60.6  \\[0.5ex]
FFTRadNet \cite{radial} & 3.8 & 74.0 \\[0.5ex]
C-M DNN \cite{radar_multitask_learning} & 7.7 & 80.4  \\[0.5ex]
\textbf{TransRadar} & \textbf{3.4} & \textbf{81.1}  \\[0.5ex]

 % remove axial unet accuracy
\hline % inserts single-line  
\end{tabular} 
\caption{Semantic segmentation results on the RADIal dataset \cite{radial}. Our method outperforms most recent state-of-the-art methods in both metrics. The best scores per column are in bold. '-' is an unreported value with no replicable results.}
\label{tab:radial_seg}
\end{table}

\begin{table}[h]  
\centering % centering table  
\setlength{\tabcolsep}{4pt}
\begin{tabular}{l c c c c} % creating 10 columns  
\hline   
Backbone  & \% AP $\uparrow$ & \% AR $\uparrow$ & R(m) $\downarrow$ & A($^{\circ}$)$\downarrow$  \\ [0.5 ex]  
\hline  
Pixor \cite{pixor}  & 96.6 & 81.7 &  0.10 & 0.20  \\[0.5ex]
FFTRadNet \cite{radial} & 96.8 & 82.2 &  \textbf{0.11} & 0.17  \\[0.5ex]
C-M DNN \cite{radar_multitask_learning} & 96.9 & 83.5 & - & - \\[0.5ex]
\textbf{TransRadar} & \textbf{97.3} & \textbf{98.4} & \textbf{0.11} & \textbf{0.10}  \\[0.5ex]

 % remove axial unet accuracy
\hline % inserts single-line  
\end{tabular} 
\caption{Object detection results on the RADIal dataset. Our method yields an increase in the average recall and a significant decrease in the angle regression error. The best scores per column are in bold. '-' is an unreported value with no replicable results.}
\label{tab:radial_od}
\end{table}

\subsection{Discussion \& Ablation Study}\label{ablation}
%\subsubsection{Different Backbone Architectures}
\noindent\textbf{Different Backbone Architectures:}
To evaluate the effect of using our loss function, we compare different other backbones using the same configuration on the CARRADA dataset. Tested backbones include available state-of-the-art methods and other transformer architectures such as ViT \cite{dosovitskiy2020vit}, UNETR \cite{UNETR}, ConViT \cite{convit}, and CSWin Transformer \cite{cswin}. This allows us to evaluate both the loss function with other state-of-the-art models and our adaptive-directional attention with other attention-based techniques. Table \ref{tab:ablation_vit} lists the quantitative comparison between them. Other than TMVA-Net, models were implemented with the same encoding and decoding as our adaptive-directional attention block. We notice that our loss improves TMVA-Net's performance significantly in both RD and RA mIoU scores. TransRadar still outperforms all other attention models and shows that the sparse nature of the adaptive-directional attention yields the best results in radar perception. To evaluate the effect of the adaptive sampling, we implement our model by applying attention to unshifted and unmodulated axes. Adding adaptive-directional sampling yields an increase of $1.40\%$ in the RD mIoU and a $4.04\%$ increase in the RA mIoU, while using less parameters than previous state-of-the-art methods. \\

\begin{table}[t]  
\small
\centering % centering table  
\begin{tabular}{l c c c c} % creating 10 columns  
\hline   
\multirow{1}{*}{Architecture} & \multicolumn{1}{c}{Param. (M)} & \multicolumn{1}{c}{mIoU$_{RD}$(\%)} & \multicolumn{1}{c}{mIoU$_{RA}$(\%)} \\
\hline  
UNETR \cite{UNETR} & 165.0 & 52.5 &  34.2 \\[0.5ex]  
CSWin \cite{cswin} & 83.0 & 25.0 &  21.9\\[0.5ex]  
ViT \cite{dosovitskiy2020vit} & 238.9 & 28.5 & 36.9 \\[0.5ex]  
% AxialDeepLab \cite{axialdeeplab} & 4.0 & 57.6 & 43.0\\  [0.5ex]
UNet \cite{unet} &  184.4 & 53.1 & 38.4 \\  [0.5ex]
TMVA-Net \cite{2021multiview} & 5.6 & 60.7 & 43.1\\  [0.5ex]
% PeakConv \cite{zhang2023peakconv} & 6.3 &  & \\  [0.5ex]
No Adaptive  & \textbf{4.0} & 61.9 & 42.3 \\  [0.5ex]
\textbf{TransRadar} & 4.8 & \textbf{63.9} & \textbf{47.5} \\  
 % remove axial unet accuracy
\hline % inserts single-line  
\end{tabular} 
\caption{Different backbones using our proposed loss configuration. The best scores are in bold. 'No Adaptive' refers to the implementation of our method where no offset or modulating to the axis sampling is introduced (i.e. straight-line rows and columns).} \label{tab:ablation_vit}
\end{table}

\begin{table}[h]
\centering
\begin{tabular}{l|c|c}
% \hline
Model & mIoU$_{RD}$ & mIoU$_{RA}$ \\
\hline
Sampling only & 62.9 & 47.4\\
Attention without sampling & 63.0 & 43.3\\
Attention with normal sampling & 64.1 & 45.7\\
\textbf{TransRadar} & 63.9 & 47.5 \\
\hline
\end{tabular}
\caption{Ablation experiment for the adaptive-directional attention head. We report segmentation performance on CARRADA dataset in terms of mIoU for the RA and RD maps.}
\label{tab:ablation_head}
\end{table}

%\subsubsection
%$\mathcal{L}_{OC} $
\begin{table}
\small
\centering
\begin{tabular}{c  c  c  c  c | c | c }
\hline
\multicolumn{5}{c|}{Loss} & \multicolumn{1}{c|}{RD} & \multicolumn{1}{c}{RA} \\
\cmidrule(lr){1-5}\cmidrule(lr){6-6}\cmidrule(lr){7-7}
  % OC & CL & SDL & CoL & MV & \multicolumn{1}{c|}{mIoU}  & \multicolumn{1}{c}{mIoU} \\
 $\mathcal{L}_{OC} $ & $\mathcal{L}_{CL} $ & $\mathcal{L}_{SD} $ & $\mathcal{L}_{CoL} $ & $\mathcal{L}_{MV} $ & \multicolumn{1}{c|}{mIoU}  & \multicolumn{1}{c}{mIoU} \\
\hline
\checkmark & \checkmark & & & \checkmark & 3.7 & 7.5 \\ 
 \checkmark & \checkmark & \checkmark & &  & 61.9 &  37.5 \\ 
& \checkmark & \checkmark &  & \checkmark & 61.2 & 45.9  \\ 
\checkmark& &\checkmark & & \checkmark & 62.3 & 42.2 \\ 
\checkmark&\checkmark &\checkmark &\checkmark & & 62.9 & 47.4 \\
 \checkmark &\checkmark  &\checkmark & &\checkmark &  \textbf{63.9} & \textbf{47.5} \\ 

\hline
\end{tabular}
\caption{Comparison of performance of loss functions. $\mathcal{L}_{OC} $ is the object centric-focal loss, $\mathcal{L}_{CL} $ is the class-agnostic object localization loss, $\mathcal{L}_{CA} $ is the sum of previous terms, $\mathcal{L}_{SD} $ is the soft Dice loss, $\mathcal{L}_{MV} $ is the multi-view range matching loss, and CoL is the coherence loss used in \cite{2021multiview}. The best scores are in bold.}\label{tab:ap_comp}
\end{table}

\noindent\textbf{Ablation for the adaptive-directional attention:}
We also perform ablation experiments on the adaptive-directional attention head. We show the semantic segmentation performance on the test split of the CARRADA dataset in Table \ref{tab:ablation_head}. Noticeably, attention contributes to the increments in RD map performance, while the directional sampling contributes to RA’s mIoU.

\noindent\textbf{Evaluation of Loss Functions:}
We further test the effect of the loss functions on the learning of our method, where we test our model under different combinations of the functions defined in Section \ref{total_loss}. Removing $\mathcal{L}_{SD} $ yields poor prediction scores, which showcases its necessity in this task. %importance and why we decided to keep it. 
Using our model without RA-RD coherence yields a poor RA score, while using a coherence loss boosts RA's score by at least 3.5\%. We also report the effects of $\mathcal{L}_{OC} $ and $\mathcal{L}_{CL} $, separately, or both combined ($\mathcal{L}_{CA} $). Removing $\mathcal{L}_{OC} $ from the $\mathcal{L}_{CA} $ term reduces RD score heavily while removing $\mathcal{L}_{CL} $ from $\mathcal{L}_{CA} $ reduces RA score. Localization is a harder task in RA maps than it is in RD due to its larger resolution which results in a more pronounced effect from $\mathcal{L}_{CL} $. Lastly, we compare the effect of introducing our $\mathcal{L}_{MV} $ loss instead of the baseline coherence loss. Following our discussion in Section \ref{total_loss}, $\mathcal{L}_{MV} $ remedies the problem of RA reducing RD's accuracy, where we notice an increase in the accuracy of RA without compromising RD scores. 
\section{Conclusion}
We introduce a novel attention-based architecture for the task of semantic segmentation on radar frequency images, named TransRadar. Our method uses an adaptive-directional attention block and a novel loss function tailored to the needs of radar perception. Our model achieves state-of-the-art performance on two semantic segmentation radar frequency datasets, CARRADA \cite{carrada} and RADIal \cite{radial}, using a smaller model size. Our proposed method also achieves improved performance for the task of object detection in radar images.\\
Paths of future works include implementing approaches that fuse radar input with RGB images to produce more robust predictions. The ability to fuse both data sources is promising in creating a new standard for automotive driving.%, where the fusion would benefit from the best tool in each modality. %We hope for our work to encourage researchers in the field to pursue further work on radar perception and bring more standardized datasets and robust models to contribute to this growing field. 

% We introduce a novel attention-based architecture for the task of semantic segmentation on radar frequency images, named TransRadar. Our method uses an adaptive-directional attention block that is associated with a novel loss function tailored for the needs of radar perception, allowing models to increase their localization abilities and decrease the background/foreground misclassification error. Our model sets new scores for two state-of-the-art semantic segmentation radar frequency datasets, CARRADA \cite{carrada} and RADIal \cite{radial}, using a smaller model size. We aim to further pursue making a generalizable model that can be fused with RGB images to produce more robust predictions. The ability to fuse both data sources is promising in creating a new standard for automotive driving, where we can use the best aspects of each acquisition tool. We hope that our work will encourage the computer vision community to pursue research in the task of radar perception and bring more standardized datasets and robust models to contribute to this growing field. 

%%%%%%%%% REFERENCES
{\small
\bibliographystyle{ieee_fullname}
\bibliography{egbib}

\begin{thebibliography}{10}\itemsep=-1pt

\bibitem{radsegnet}
Kshitiz Bansal, Keshav Rungta, and Dinesh Bharadia.
\newblock Radsegnet: A reliable approach to radar camera fusion.
\newblock {\em arXiv preprint arXiv:2208.03849}, 2022.

\bibitem{robotcar}
Dan Barnes, Matthew Gadd, Paul Murcutt, Paul Newman, and Ingmar Posner.
\newblock The oxford radar robotcar dataset: A radar extension to the oxford
  robotcar dataset.
\newblock In {\em Proceedings of the IEEE International Conference on Robotics
  and Automation (ICRA)}, Paris, 2020.

\bibitem{aspp}
Liang-Chieh Chen, George Papandreou, Iasonas Kokkinos, Kevin Murphy, and Alan~L
  Yuille.
\newblock Deeplab: Semantic image segmentation with deep convolutional nets,
  atrous convolution, and fully connected crfs.
\newblock {\em IEEE transactions on pattern analysis and machine intelligence},
  40(4):834--848, 2017.

\bibitem{deeplabv3+}
Liang-Chieh Chen, Yukun Zhu, George Papandreou, Florian Schroff, and Hartwig
  Adam.
\newblock Encoder-decoder with atrous separable convolution for semantic image
  segmentation.
\newblock In {\em Proceedings of the European conference on computer vision
  (ECCV)}, pages 801--818, 2018.

\bibitem{cswin}
Xiaoyi Dong, Jianmin Bao, Dongdong Chen, Weiming Zhang, Nenghai Yu, Lu Yuan,
  Dong Chen, and Baining Guo.
\newblock Cswin transformer: A general vision transformer backbone with
  cross-shaped windows.
\newblock In {\em Proceedings of the IEEE/CVF Conference on Computer Vision and
  Pattern Recognition}, pages 12124--12134, 2022.

\bibitem{dosovitskiy2020vit}
Alexey Dosovitskiy, Lucas Beyer, Alexander Kolesnikov, Dirk Weissenborn,
  Xiaohua Zhai, Thomas Unterthiner, Mostafa Dehghani, Matthias Minderer, Georg
  Heigold, Sylvain Gelly, Jakob Uszkoreit, and Neil Houlsby.
\newblock An image is worth 16x16 words: Transformers for image recognition at
  scale.
\newblock {\em ICLR}, 2021.

\bibitem{9000872}
Di Feng, Christian Haase-Schütz, Lars Rosenbaum, Heinz Hertlein, Claudius
  Gläser, Fabian Timm, Werner Wiesbeck, and Klaus Dietmayer.
\newblock Deep multi-modal object detection and semantic segmentation for
  autonomous driving: Datasets, methods, and challenges.
\newblock {\em IEEE Transactions on Intelligent Transportation Systems},
  22(3):1341--1360, 2021.

\bibitem{9048939}
Xiangyu Gao, Guanbin Xing, Sumit Roy, and Hui Liu.
\newblock Experiments with mmwave automotive radar test-bed.
\newblock In {\em 2019 53rd Asilomar Conference on Signals, Systems, and
  Computers}, pages 1--6, 2019.

\bibitem{rampcnn}
Xiangyu Gao, Guanbin Xing, Sumit Roy, and Hui Liu.
\newblock Ramp-cnn: A novel neural network for enhanced automotive radar object
  recognition.
\newblock {\em IEEE Sensors Journal}, 21(4):5119--5132, 2021.

\bibitem{UNETR}
Ali Hatamizadeh, Yucheng Tang, Vishwesh Nath, Dong Yang, Andriy Myronenko,
  Bennett Landman, Holger~R Roth, and Daguang Xu.
\newblock Unetr: Transformers for 3d medical image segmentation.
\newblock In {\em Proceedings of the IEEE/CVF winter conference on applications
  of computer vision}, pages 574--584, 2022.

\bibitem{axial}
Jonathan Ho, Nal Kalchbrenner, Dirk Weissenborn, and Tim Salimans.
\newblock Axial attention in multidimensional transformers.
\newblock {\em arXiv preprint arXiv:1912.12180}, 2019.

\bibitem{ccnet}
Zilong Huang, Xinggang Wang, Lichao Huang, Chang Huang, Yunchao Wei, and Wenyu
  Liu.
\newblock Ccnet: Criss-cross attention for semantic segmentation.
\newblock In {\em Proceedings of the IEEE/CVF International Conference on
  Computer Vision}, pages 603--612, 2019.

\bibitem{huber}
Peter~J. Huber.
\newblock {Robust Estimation of a Location Parameter}.
\newblock {\em The Annals of Mathematical Statistics}, 35(1):73 -- 101, 1964.

\bibitem{TRODNet}
Tiezhen Jiang, Long Zhuang, Qi An, Jianhua Wang, Kai Xiao, and Anqi Wang.
\newblock T-rodnet: Transformer for vehicular millimeter-wave radar object
  detection.
\newblock {\em IEEE Transactions on Instrumentation and Measurement}, 72:1--12,
  2023.

\bibitem{radar_multitask_learning}
Yi Jin, Anastasios Deligiannis, Juan-Carlos Fuentes-Michel, and Martin Vossiek.
\newblock Cross-modal supervision-based multitask learning with automotive
  radar raw data.
\newblock {\em IEEE Transactions on Intelligent Vehicles}, pages 1--15, 2023.

\bibitem{wce}
Prannay Kaul, Daniele de Martini, Matthew Gadd, and Paul Newman.
\newblock Rss-net: Weakly-supervised multi-class semantic segmentation with
  fmcw radar.
\newblock In {\em 2020 IEEE Intelligent Vehicles Symposium (IV)}, pages
  431--436, 2020.

\bibitem{adam}
Diederik~P. Kingma and Jimmy Ba.
\newblock Adam: {A} method for stochastic optimization.
\newblock In Yoshua Bengio and Yann LeCun, editors, {\em 3rd International
  Conference on Learning Representations, {ICLR} 2015, San Diego, CA, USA, May
  7-9, 2015, Conference Track Proceedings}, 2015.

\bibitem{feature_pyramid_object_detector}
Tsung-Yi Lin, Piotr Doll{\'a}r, Ross Girshick, Kaiming He, Bharath Hariharan,
  and Serge Belongie.
\newblock Feature pyramid networks for object detection.
\newblock In {\em Proceedings of the IEEE conference on computer vision and
  pattern recognition}, pages 2117--2125, 2017.

\bibitem{focal_loss_object_detector}
Tsung-Yi Lin, Priya Goyal, Ross Girshick, Kaiming He, and Piotr Dollár.
\newblock Focal loss for dense object detection.
\newblock {\em IEEE Transactions on Pattern Analysis and Machine Intelligence},
  42(2):318--327, 2020.

\bibitem{liu2021Swin}
Ze Liu, Yutong Lin, Yue Cao, Han Hu, Yixuan Wei, Zheng Zhang, Stephen Lin, and
  Baining Guo.
\newblock Swin transformer: Hierarchical vision transformer using shifted
  windows.
\newblock In {\em Proceedings of the IEEE/CVF International Conference on
  Computer Vision (ICCV)}, 2021.

\bibitem{fcn8}
Jonathan Long, Evan Shelhamer, and Trevor Darrell.
\newblock Fully convolutional networks for semantic segmentation.
\newblock In {\em Proceedings of the IEEE conference on computer vision and
  pattern recognition}, pages 3431--3440, 2015.

\bibitem{8904734}
Michael Meyer and Georg Kuschk.
\newblock Automotive radar dataset for deep learning based 3d object detection.
\newblock In {\em 2019 16th European Radar Conference (EuRAD)}, pages 129--132,
  2019.

\bibitem{9299052}
Farzan~Erlik Nowruzi, Dhanvin Kolhatkar, Prince Kapoor, Fahed Al~Hassanat,
  Elnaz~Jahani Heravi, Robert Laganiere, Julien Rebut, and Waqas Malik.
\newblock Deep open space segmentation using automotive radar.
\newblock In {\em 2020 IEEE MTT-S International Conference on Microwaves for
  Intelligent Mobility (ICMIM)}, pages 1--4, 2020.

\bibitem{polarnet}
Farzan~Erlik Nowruzi, Dhanvin Kolhatkar, Prince Kapoor, Elnaz~Jahani Heravi,
  Fahed~Al Hassanat, Robert Lagani{\`{e}}re, Julien Rebut, and Waqas Malik.
\newblock Polarnet: Accelerated deep open space segmentation using automotive
  radar in polar domain.
\newblock {\em CoRR}, abs/2103.03387, 2021.

\bibitem{2021multiview}
Arthur Ouaknine, Alasdair Newson, Patrick P{\'e}rez, Florence Tupin, and Julien
  Rebut.
\newblock Multi-view radar semantic segmentation.
\newblock In {\em Proceedings of the IEEE/CVF International Conference on
  Computer Vision}, pages 15671--15680, 2021.

\bibitem{carrada}
Arthur Ouaknine, Alasdair Newson, Julien Rebut, Florence Tupin, and Patrick
  Pérez.
\newblock Carrada dataset: Camera and automotive radar with range- angle-
  doppler annotations.
\newblock In {\em 2020 25th International Conference on Pattern Recognition
  (ICPR)}, pages 5068--5075, 2021.

\bibitem{intro1}
Andras Palffy, Jiaao Dong, Julian~FP Kooij, and Dariu~M Gavrila.
\newblock Cnn based road user detection using the 3d radar cube.
\newblock {\em IEEE Robotics and Automation Letters}, 5(2):1263--1270, 2020.

\bibitem{radial}
Julien Rebut, Arthur Ouaknine, Waqas Malik, and Patrick P{\'e}rez.
\newblock Raw high-definition radar for multi-task learning.
\newblock In {\em Proceedings of the IEEE/CVF Conference on Computer Vision and
  Pattern Recognition}, pages 17021--17030, 2022.

\bibitem{unet}
O. Ronneberger, P.Fischer, and T. Brox.
\newblock U-net: Convolutional networks for biomedical image segmentation.
\newblock In {\em Medical Image Computing and Computer-Assisted Intervention
  (MICCAI)}, volume 9351 of {\em LNCS}, pages 234--241. Springer, 2015.
\newblock (available on arXiv:1505.04597 [cs.CV]).

\bibitem{radar_scenes}
Ole Schumann, Markus Hahn, Nicolas Scheiner, Fabio Weishaupt, Julius~F Tilly,
  J{\"u}rgen Dickmann, and Christian W{\"o}hler.
\newblock Radarscenes: A real-world radar point cloud data set for automotive
  applications.
\newblock In {\em 2021 IEEE 24th International Conference on Information Fusion
  (FUSION)}, pages 1--8. IEEE, 2021.

\bibitem{region_based_object_detector}
Abhinav Shrivastava, Abhinav Gupta, and Ross Girshick.
\newblock Training region-based object detectors with online hard example
  mining.
\newblock In {\em Proceedings of the IEEE conference on computer vision and
  pattern recognition}, pages 761--769, 2016.

\bibitem{axialdeeplab}
Huiyu Wang, Yukun Zhu, Bradley Green, Hartwig Adam, Alan Yuille, and
  Liang-Chieh Chen.
\newblock Axial-deeplab: Stand-alone axial-attention for panoptic segmentation.
\newblock In {\em Computer Vision--ECCV 2020: 16th European Conference,
  Glasgow, UK, August 23--28, 2020, Proceedings, Part IV}, pages 108--126.
  Springer, 2020.

\bibitem{rodnet}
Yizhou Wang, Zhongyu Jiang, Xiangyu Gao, Jenq-Neng Hwang, Guanbin Xing, and Hui
  Liu.
\newblock Rodnet: Radar object detection using cross-modal supervision.
\newblock In {\em 2021 IEEE Winter Conference on Applications of Computer
  Vision (WACV)}, pages 504--513, 2021.

\bibitem{convit}
Haiping Wu, Bin Xiao, Noel Codella, Mengchen Liu, Xiyang Dai, Lu Yuan, and Lei
  Zhang.
\newblock Cvt: Introducing convolutions to vision transformers.
\newblock In {\em Proceedings of the IEEE/CVF International Conference on
  Computer Vision}, pages 22--31, 2021.

\bibitem{xu2022sparse}
Shuangjie Xu, Rui Wan, Maosheng Ye, Xiaoyi Zou, and Tongyi Cao.
\newblock Sparse cross-scale attention network for efficient lidar panoptic
  segmentation.
\newblock In {\em Proceedings of the AAAI Conference on Artificial
  Intelligence}, volume~36, pages 2920--2928, 2022.

\bibitem{xue2022efficient}
Ruixiang Xue, Jianqiang Wang, and Zhan Ma.
\newblock Efficient lidar point cloud geometry compression through neighborhood
  point attention.
\newblock {\em arXiv preprint arXiv:2208.12573}, 2022.

\bibitem{pixor}
Bin Yang, Wenjie Luo, and Raquel Urtasun.
\newblock Pixor: Real-time 3d object detection from point clouds.
\newblock In {\em Proceedings of the IEEE conference on Computer Vision and
  Pattern Recognition}, pages 7652--7660, 2018.

\bibitem{unified_focal_loss}
Michael Yeung, Evis Sala, Carola-Bibiane Schönlieb, and Leonardo Rundo.
\newblock Unified focal loss: Generalising dice and cross entropy-based losses
  to handle class imbalanced medical image segmentation.
\newblock {\em Computerized Medical Imaging and Graphics}, 95:102026, 2022.

\bibitem{raddet}
Ao Zhang, Farzan~Erlik Nowruzi, and Robert Laganiere.
\newblock Raddet: Range-azimuth-doppler based radar object detection for
  dynamic road users.
\newblock In {\em 2021 18th Conference on Robots and Vision (CRV)}, pages
  95--102, 2021.

\bibitem{zhang2021sparse}
Biao Zhang, Ivan Titov, and Rico Sennrich.
\newblock Sparse attention with linear units.
\newblock {\em arXiv preprint arXiv:2104.07012}, 2021.

\bibitem{hand-gesture4}
Guoqiang Zhang, Haopeng Li, and Fabian Wenger.
\newblock Object detection and 3d estimation via an fmcw radar using a fully
  convolutional network.
\newblock In {\em ICASSP 2020-2020 IEEE International Conference on Acoustics,
  Speech and Signal Processing (ICASSP)}, pages 4487--4491. IEEE, 2020.

\bibitem{zhang2023peakconv}
Liwen Zhang, Xinyan Zhang, Youcheng Zhang, Yufei Guo, Yuanpei Chen, Xuhui
  Huang, and Zhe Ma.
\newblock Peakconv: Learning peak receptive field for radar semantic
  segmentation.
\newblock In {\em Proceedings of the IEEE/CVF Conference on Computer Vision and
  Pattern Recognition}, pages 17577--17586, 2023.

\bibitem{hand-gesture2}
Zhenyuan Zhang, Zengshan Tian, Ying Zhang, Mu Zhou, and Bang Wang.
\newblock u-deephand: Fmcw radar-based unsupervised hand gesture feature
  learning using deep convolutional auto-encoder network.
\newblock {\em IEEE Sensors Journal}, 19(16):6811--6821, 2019.

\bibitem{hand-gesture3}
Zhenyuan Zhang, Zengshan Tian, and Mu Zhou.
\newblock Latern: Dynamic continuous hand gesture recognition using fmcw radar
  sensor.
\newblock {\em IEEE Sensors Journal}, 18(8):3278--3289, 2018.

\end{thebibliography}
}

\end{document}